# Dual-Forward Path Teacher Knowledge Distillation: Bridging The Capacity Gap Between Teacher and Student

Tong Li, Long Liu, *Member, IEEE*, Yihang Hu, Hu Chen, and Shifeng Chen

*Abstract*—Knowledge distillation (KD) provides an effective way to improve the performance of a student network under the guidance of pre-trained teachers. However, this approach usually brings in a large capacity gap between teacher and student networks, limiting the distillation gains. Previous methods addressing this problem either discard accurate knowledge representation or fail to dynamically adjust the transferred knowledge, which is less effective in addressing the capacity gap problem and hinders students from achieving comparable performance with the pre-trained teacher. In this work, we extend the ideology of prompt-based learning to address the capacity gap problem, and propose Dual-Forward Path Teacher Knowledge Distillation (DFPT-KD), which replaces the pre-trained teacher with a novel dual-forward path teacher to supervise the learning of student. The key to DFPT-KD is prompt-based tuning, *i.e.*, establishing an additional prompt-based forward path within the pre-trained teacher and optimizing it with the pre-trained teacher frozen to make the transferred knowledge compatible with the representation ability of the student. Extensive experiments demonstrate that DFPT-KD leads to trained students performing better than the vanilla KD. To make the transferred knowledge better compatible with the representation abilities of the student, we further fine-tune the whole prompt-based forward path, yielding a novel distillation approach dubbed DFPT-KDt. By extensive experiments, it is shown that DFPT-KDt improves upon DFPT-KD and achieves state-of-the-art accuracy performance.

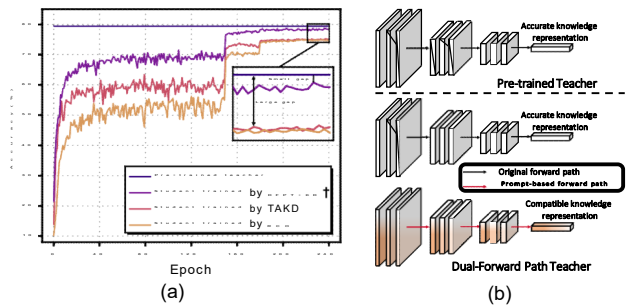

Fig. 1. (a) Unlike existing adaptive KD methods, our approach effectively bridges the capacity gap and promotes the student to achieve more comparable performance with the pre-trained teacher. (b) Compared to the pre-trained teacher, which only generates accurate knowledge representation, the dual-forward path teacher can simultaneously generate accurate and compatible knowledge representations for the student.

## I. INTRODUCTION

LARGE deep neural networks have made remarkable success in numerous computer vision tasks like image classification [1]–[3], object detection [4], and semantic segmentation [5], [6]. However, these high-performance networks are always accompanied by enormous parameters and computation costs, rendering them hard to deploy on embedded devices with limited computational and storage resources. Knowledge distillation (KD) [7] provides a promising solution to train lightweight models by transferring knowledge from high-capacity teacher models (called teachers for short) to low-capacity student ones (called students for short), garnering increasing attention from the community. In the past several years, various KD methods [8]–[19] have been proposed, which follow an assumption that well-performing teachers contribute to better distillation performance and focus on how to guide students to learn the knowledge of pre-trained teachers.

However, some studies have found that when the gap in capacity between teachers and students is large, the distillation effect may be negatively affected. Cho et al. [20] discovered that the full-trained teacher might not be the optimal choice for distillation. Mirzadeh et al. [21] experimentally verified that the pre-trained, well-performing teacher is unsuitable for distilling a high-performance student. This phenomenon is known as the capacity gap problem, which limits the distillation gains. The underlying cause of this problem, as explained in [22] and [9], is that during the pre-training of teachers, the confidence in the target class is constantly increased to minimize the negative log-likelihood. In contrast, the confidence in non-target classes is significantly suppressed. In this case, the knowledge of pre-trained teachers may not be conducive to the learning of compact students with limited representation abilities.

In the past years, several works, also known as adaptive KD methods, have been proposed to address the capacity gap problem. Most of these are multi-stage approaches [21], [23]–[27], which introduce one or more teacher assistants (TAs) to gradually bridge the capacity gap with a philosophy of decomposing the single difficult distillation task into multiple easier ones. On the other hand, one-stage methods [20], [28]–[30] utilize knowledge distilled from the teacher training footsteps to accomplish knowledge transferring, making the learning process of students more smooth and gradual. Although decent performances have been achieved, as shown in Fig. 1 (a), these methods still struggle to achieve further improvement due to two limitations : (1) Either multi-stage or one-stage methods

This work is supported by the National Natural Science Foundation of China under Grant No. 61673318, the Xi'an Science and Technology Project under Grant NO. 22GXFW0096, and the Key Industry Chain Project of Shaanxi Province under Grant NO. 2020ZDLGY04-04. *(Corresponding author: Long Liu.)*

Tong Li, Long Liu, Yihang Hu, Hu Chen, and Shifeng Chen are with the School of Automation and Information Engineering, Xi'an University of Technology, Xi'an 710048, China (e-mail: tli@stu.xaut.edu.cn; liulong@xaut.edu.cn; 2230320140@stu.xaut.edu.cn; 2220321223@stu.xaut.edu.cn; 2240320145@stu.xaut.edu.cn).

make the transferred knowledge easier for the student to learn by decreasing the performance of teachers, resulting in the distillation process lacking accurate knowledge representation in most phases. That hinders the student from achieving comparable performance with the pre-trained teacher. (2) Existing solutions simply decrease teachers' capacity or modify parameter space to adjust their knowledge, failing to dynamically adjust the transferred knowledge to be compatible with the varying representation abilities of the student at each distillation phase. Consequently, they still provide the student with general knowledge, which is less effective in alleviating the capacity gap problem. Unfortunately, teacher models typically have only one forward path, making them unable to provide the student with accurate and compatible knowledge representations simultaneously. Therefore, a question arises: Can we dynamically adjust the transferred knowledge without discarding the accurate knowledge representation and simultaneously provide the student with these two kinds of knowledge?

In the natural language processing (NLP) and computer vision (CV) communities, prompt-based learning has attracted increased attention because of its astounding transfer performance on various downstream tasks. Typically, the prompt is a task-relevant description added to the downstream input, guiding or directing the downstream task to the pre-trained model. Their key idea is reformulating the downstream task with an appropriate prompt design, making it close to the task that the original pre-trained model solved, rather than fine-tuning pre-trained models to adapt to downstream tasks. In other words, they did not decrease the performance of the pre-trained model on the original task. Following this ideology, various multi-modal and single-modal prompt-based tuning paradigms have been proposed, such as CoCoOp [31], MaPLe [32], VPT [33], and LION [34]. Though achieving remarkable success, these approaches are tailored for various NLP or CV downstream tasks, inapplicably addressing the capacity gap problem in knowledge distillation.

In this paper, we extend the ideology of prompt-based learning and propose **D**ual-**F**orward **P**ath **T**eacher **K**nowledge **D**istillation (DFPT-KD). Different from classical KD methods, as shown in Fig. 1 (b), DFPT-KD replace the pre-trained teacher with a novel dual-forward path teacher to provide the student with accurate and compatible knowledge representations simultaneously. The key to DFPT-KD is prompt-based tuning, *i.e.*, establishing an additional prompt-based forward path within the pre-trained teacher and optimizing it to make the transferred knowledge compatible with the representation abilities of the student. Specifically, DFPT-KD constructs a set of prompt blocks to generate student-specific discriminative knowledge prompts for training inputs. By fusing the intermediate feature of each input with the corresponding learned prompts, DFPT-KD can establish a prompt-based forward path within the pre-trained teacher. During distillation, we optimize this forward path by only tuning prompt blocks through bidirectional supervision from both the teacher's original forward path (equivalent to the pre-trained teacher) and the student. In this way, we can make the transferred knowledge follow an easy-to-hard principle and be highly compatible with the representation abilities of the student while still leveraging the original forward path to provide accurate knowledge representation for the student. In addition, to make the transferred knowledge better compatible with the representation abilities of the student, we further fine-tune the whole prompt-based forward path without discarding the accurate knowledge representation, yielding a novel distillation approach dubbed DFPT-KDt. DFPT-KDt improves upon DFPT-KD, boosting the performance of our approach to a higher level.

Extensive experiments on CIFAR-100 [35], ImageNet [36], and CUB-200 [37] datasets show that DFPT-KD leads to the trained student with better performance than vanilla KD, achieving comparable or even better performance with other state-of-the-art distillation methods. Moreover, DFPT-KDt performs better than DFPT-KD and achieves new state-of-the-art performance. For instance, DFPT-KDt obtains 78.29% Top-1 accuracy with ShuffleNetV1 on the CIFAR-100, surpassing teacher WRN-40-2 by 2.68%, verifying the superiority of our approach.

In a nutshell, we made the following main contributions:

- To the best of our knowledge, we are the first to utilize the prompt-based tuning strategy addressing the capacity gap problem in knowledge distillation.
- We establish an additional prompt-based forward path within the pre-trained teacher and optimize it through bidirectional supervision to make their knowledge follow an easy-to-hard principle and be compatible with the representation abilities of the student.
- Extensive experiments on CIFAR-100, ImageNet, and CUB-200 datasets are conducted to showcase that our approach consistently outperforms existing KD methods, achieving new state-of-the-art performance.

## II. Related Work

### A. Knowledge Distillation

The concept of knowledge distillation was proposed by Hinton et al. [7], where a smaller student model tries to mimic the predicted distribution of a pre-trained large teacher model by minimizing the Kullback-Leibler (KL) divergence loss with a temperature factor. Since then, various distillation methods have been proposed, which, based on the knowledge representation, can be roughly categorized into two types: logits-based distillation [8], [10]–[12], [17], [38]–[41] and feature-based distillation [13]–[16], [19], [42]–[48]. These methods rely heavily on a well-performing pre-trained teacher model. However, the large capacity gap between teachers and students limits the distillation gains.

To alleviate the capacity gap problem, adaptive KD methods have been proposed. One type of solution is the multi-stage approach [21], [23]–[27]. TAKD [21] bridged this gap by training intermediate-sized teacher assistants (TAs). DGKD [23] used a densely guiding manner to train each TA with higher TAs and the teacher to alleviate error avalanche problems in TAKD. ResKD [24] used additional residual networks to bridge the capacity gap. AAKD [27] introduced a knowledge sample selection strategy and an adaptive teacher strategy to automatically select suitable samples and teachers. Another type of solution is the one-stage approach [20], [28]–[30], [49]. ESKD [20] proposed an early stopping strategy for the teacher, facilitating a more favourable solution. RCO [28] constructed a gradually mimicking sequence by selecting some checkpoints from the training footpath to guide the student. Pro-KD [29] distilled knowledge from the teacher training routes, providing a smoother training path for the student. EKD [30] used an evolutionary teacher to supervise the student on-the-fly. CKD [49] utilized the student's participation to assist the teacher in transferring appropriate knowledge. However, these methods either make the distillation process lack accurate knowledge representation or can not dynamically adjust the transferred knowledge to be compatible with the representation abilities of the student, which is less effective in addressing the capacity gap problem and hinders the student from achieving comparable performance with the pre-trained teacher.

*B. Prompt-Based Learning*

This topic originates from the NLP community and is popularized by BERT [50] and GPT [51] series. It utilizes task-relevant descriptions added to the downstream input to enhance the understanding of large language models for downstream tasks rather than fine-tuning the pre-trained model. Subsequently, numerous studies focused on devising effective prompt strategies for extracting knowledge from pre-trained models. For instance, CoOp [52] and CoCoOp [31] replace the hand-crafted hard prompts (*e.g.,* "a photo of a {class}") with learnable text continuous tokens (also known as soft prompts) for fine-tuning with the pre-trained model frozen. VPT [33] moved the learnable continuous tokens from the text encoder to the image one and proposed to insert the "visual soft prompt" to the patch sequence of an input image for fine-tuning. MaPLe [32] appended the soft prompt to the hidden representations at each layer in both the image and text encoder. LION [53] proposed to insert two equilibrium implicit layers in two ends of the pre-trained backbone with parameters frozen. In addition, prompt-based learning was also applied to various visual downstream tasks, such as object detection [54], semantic segmentation [55], and video recognition [56]. In knowledge distillation, FreeKD [57] proposed a semantic frequency prompt plugged into the pre-trained teacher for distillation on dense prediction tasks. CLIP-KD [58] aimed to distill small CLIP [59] models via relation, feature, gradient and contrastive distillation strategies. PromptKD [60] proposed a two-stage unsupervised prompt distillation framework for vision-language models, which distills a lightweight CLIP model from a large CLIP model through prompt imitation. In this work, we aim to address the capacity gap problem in knowledge distillation via the prompt-based tuning strategy.

## III. METHODOLOGY

In this section, we first retrieve the formulation of vanilla KD. Through an in-depth analysis of the mechanism of vanilla KD, we can understand why the knowledge distilled from the pre-trained teacher is not conducive to the learning of students, causing the capacity gap problem. Then, we introduce the proposed DFPT-KD and its variant DFPT-KD† in detail. The overview of the proposed method is shown in Fig. 2.

*A. Revisit of Vanilla KD*

For a C-way classification task, the logit output of the network on a training input $(x, y)$ is denoted as $z = [z_1, z_2, \ldots, z_t, \ldots z_c] \in R^{1 \times c}$, where $z_i$ denotes the logit output in the i-th class. Then, each element in softening probability distribution $p = [p_1, p_2, \ldots, p_t, \ldots p_c] \in R^{1 \times c}$ can be calculated by a softmax function:

$$p_j = \frac{e^{z_j/\tau}}{\sum_{i=1}^{c} e^{z_i/\tau}}, \quad (1)$$

where $p_j$ is the softening probability on the j-th class, and $\tau$ is a temperature factor to smooth output logits. Following DKD [9], we divide the softening prediction $p$ into predictions relevant and irrelevant to the target class. Specifically, letting $b = [p_t, p_{\setminus t}] \in R^{1 \times 2}$ represents binary probabilities of the target class and all other non-target classes:

$$p_t = \frac{e^{z_t/\tau}}{\sum_{i=1}^{c} e^{z_i/\tau}}, \quad p_{\setminus t} = \frac{\sum_{i=1, i \neq t}^{c} e^{z_i/\tau}}{\sum_{i=1}^{c} e^{z_i/\tau}}. \quad (2)$$

Meanwhile, letting $q = [q_1, q_2, \ldots, q_{t-1}, q_{t+1}, \ldots q_c] \in R^{1 \times (c-1)}$ represents probability distributions among non-target classes, each element in $q$ is calculated by:

$$q_j = \frac{e^{z_j/\tau}}{\sum_{i=1, i \neq t}^{c} e^{z_i/\tau}}. \quad (3)$$

Then, we can rewrite the distillation loss in vanilla KD as:

$$L_{KD}(p^T \parallel p^S) = p_t^T \log \frac{p_t^T}{p_t^S} + \sum_{i=1, i \neq t}^{c} p_i^T \log \frac{p_i^T}{p_i^S}, \quad (4)$$

where $L_{KD}$ represents KL divergence loss, $T$ and $S$ denote the pre-trained teacher and the student, respectively. According to Eq. (1), (2), and (3), we have $p_j = q_j \cdot p_{\setminus t}$ and $p_t + p_{\setminus t} = 1$, so we can rewrite Eq. (4) as:

$$L_{KD}(p^T \parallel p^S) = p_t^T \log \frac{p_t^T}{p_t^S} + p_{\setminus t}^T \sum_{i=1, i \neq t}^{c} q_i^T \log \frac{q_i^T \cdot p_{\setminus t}^T}{q_i^S \cdot p_{\setminus t}^S}$$

$$= p_t^T \log \frac{p_t^T}{p_t^S} + p_{\setminus t}^T \sum_{i=1, i \neq t}^{c} q_i^T \log \frac{q_i^T}{q_i^S} + q_i^T \log \frac{p_{\setminus t}^T}{p_{\setminus t}^S}$$

$$= p_t^T \log \frac{p_t^T}{p_t^S} + p_{\setminus t}^T \log \frac{p_{\setminus t}^T}{p_{\setminus t}^S} + p_{\setminus t}^T \sum_{i=1, i \neq t}^{c} q_i^T \log \frac{q_i^T}{q_i^S}. \quad (5)$$

We let:

$$L_{KD}(b^T \parallel b^S) = p_t^T \log \frac{p_t^T}{p_t^S} + p_{\setminus t}^T \log \frac{p_{\setminus t}^T}{p_{\setminus t}^S}, \quad (6)$$

$$L_{KD}(q^T \parallel q^S) = \sum_{i=1, i \neq t}^{c} q_i^T \log \frac{q_i^T}{q_i^S}. \quad (7)$$

Therefore, Eq. (5) can be rewritten as:

$$L_{KD}(p^T \parallel p^S) = L_{KD}(b^T \parallel b^S) + (1 - p_t^T) L_{KD}(q^T \parallel q^S). \quad (8)$$

As shown in Eq. (8), the vanilla KD loss is reformulated into a weighted sum of two terms. $L_{KD}(b^T \parallel b^S)$ represents the distance between the pre-trained teacher's and student's binary probabilities of the target class. $L_{KD}(q^T \| q^S)$ represents the distance between the pre-trained teacher's and student's probability distributions among non-target classes. During distillation, the pre-trained teacher has high confidence in the target class and limited confidence among non-target ones. In contrast, due to fewer parameters and simpler architecture, the student lacks confidence in the target class and spreads its confidence across non-target classes in most distillation phases. This disparity results in high values of $L_{KD}(b^T \| b^S)$ and $L_{KD}(q^T \| q^S)$, as well as a suppression for $1 - p_t^T$, resulting in the capacity gap problem.

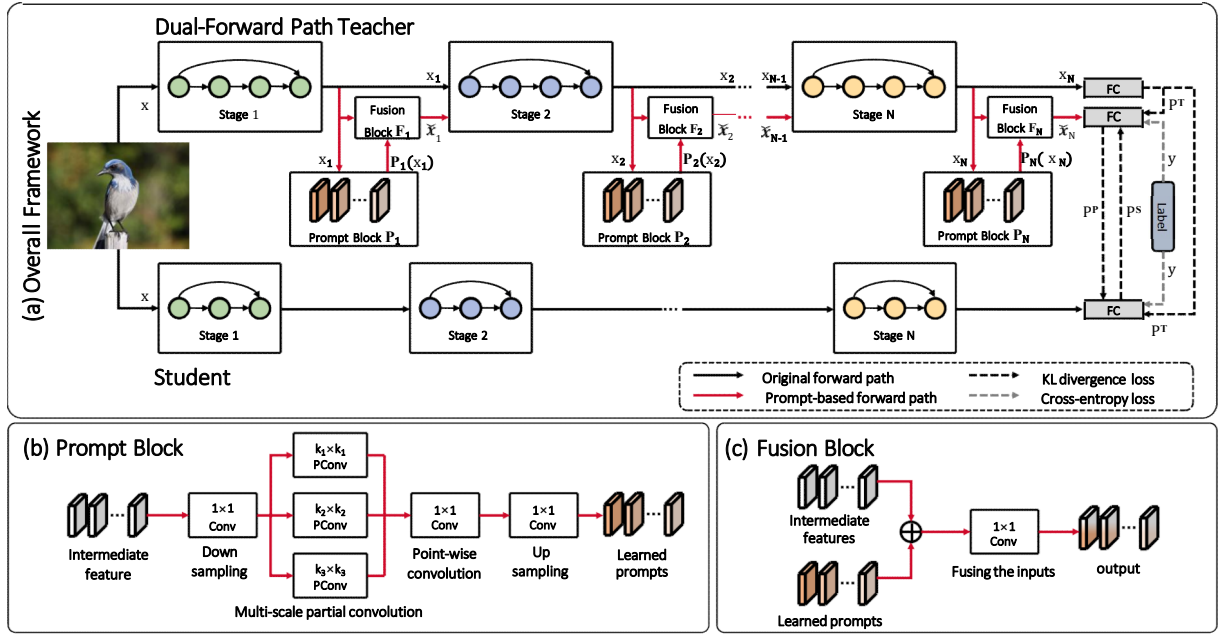

Fig. 2. An illustrative diagram of the proposed method. The upper is the overall framework, and the bottom is the detailed architecture of the prompt block and fusion block. We divide the teacher and student networks into N stages; between each stage of the teacher, a set of prompt blocks and fusion blocks is inserted to establish an additional prompt-based forward path. During distillation, this path is optimized through bidirectional supervision from the original forward path and the student, making the transferred knowledge follow an easy-to-hard principle and be compatible with the representation ability of the student without discarding accurate knowledge representation.

### B. The Dual-Forward Path Teacher

*1) Basic Primitives:* We establish an additional prompt-based forward path within the pre-trained teacher to dynamically adjust the transferred knowledge without discarding accurate knowledge representation. Technically, we introduce lightweight prompt blocks to learn student-specific knowledge prompts for each training input. In principle, these prompt blocks should be parameter efficient while having a strong representation ability to adjust the transferred knowledge. These prompt blocks should also be suitable for various network architectures, making our approach applicable to various teacher-student pairs and achieving significant improvements.

Based on the above analysis, we design lightweight prompt blocks with commonly used operation primitives that support well-optimized cuDNN functions. Specifically, we introduce a prompt block with following sequential convolution layers: $1 \times 1$ down sampling convolution, $[k_1 \times k_1, k_2 \times k_2, k_3 \times k_3]$ multi-scale partial convolution, $1 \times 1$ point-wise convolution, and $1 \times 1$ up sampling convolution. The first $1 \times 1$ down sampling convolution is used to perform channel down-sampling (the sampling rate is defined as $r_1$), reducing the computation complexity of the next operation. $[k_1 \times k_1, k_2 \times k_2, k_3 \times k_3]$ multi-scale partial convolution (the partial ratio is defined as $r_2$) provides enough spatial representation ability. A $1 \times 1$ point-wise convolution follows each partial convolution, allowing feature information to flow through all channels. Compared with conventional convolution operations, combining multi-scale partial convolution and point-wise convolution adjusts a subset of channels of the intermediate feature at a multi-scale level, saving computation costs. The last $1 \times 1$ up sampling convolution is used to perform channel up-sampling, making the magnitude of knowledge prompts the same as the original features. In addition, we introduce a fusion block to fuse the original features and corresponding prompts, encouraging an adaptive balance between these two terms.

*2) Prompt-Based Forward Path:* As pointed out in [61]–[63], the imitation ability of a network relies on visual feature representations at different levels, especially for low-level and mid-level representations. To sufficiently utilize feature information from different levels, we construct multiple stage-wise prompt blocks, which can be inserted into a given pre-trained teacher model for adjusting their knowledge. In principle, vision models typically split network architecture into multiple stages. To ensure the generality of the proposed method, we adopt the original stage partitioning approach of pre-trained teachers and incorporate the prompt blocks into their multiple stages. Specifically, a pre-trained teacher **T** consisted of a N-stage backbone $\mathbf{B} = \{\mathbf{A}_i\}_{i=1}^{N}$ and a classification head **H**, is denoted by $\mathsf{T} = \mathsf{H} \circ \mathsf{B}$. Given the j-th training sample $\mathsf{x}^j$ from a benchmark dataset $\mathcal{D} = \{(\mathsf{x}^j, \mathsf{y}^j)\}_{j=1}^{|\mathcal{D}|}$, we define the output feature of intermediate stages as $\{\mathsf{x}_1^j, \mathsf{x}_2^j, \cdots, \mathsf{x}_N^j\}$. The prompt block appended to the stage $\mathsf{A}_i$ is denoted by $\mathsf{P}_i$, then we fuse the intermediate feature $\mathsf{x}_i^j$ of the stage $\mathsf{A}_i$ and the learned knowledge prompt $\mathsf{P}_i(\mathsf{x}_i^j)$ via the fusion block to obtain the fusing prompt representation $\check{\mathsf{x}}_i^j$. This process can be formulated as

$$\check{\mathsf{x}}_i^j = \mathsf{F}_i(\mathsf{x}_i^j + \mathsf{P}_i(\mathsf{x}_i^j)), \quad i = 1, \cdots, N \quad (9)$$
$$\mathsf{x}_{i+1}^j = \mathsf{A}_{i+1}(\check{\mathsf{x}}_i^j), \quad i = 1, \cdots, N-1 \quad (10)$$

where $\mathsf{F}_i$ represents the corresponding fusion block, which consists of a set of $1 \times 1$ convolution operations.

In this way, we can establish the prompt-based forward path within the pre-trained teacher. The whole computation process of this path can be described through the following steps: 1) feeding training input images into the dual-forward path teacher; 2) generating student-specific knowledge prompts by multiple stage-wise prompt blocks; 3) fusing the intermediate features with corresponding learned prompts as input to the next stage; 4) adding an initialized classification head to top layers of the backbone for the output prediction.

*3) Bidirectional Supervision:* In DFPT-KD, the pre-trained teacher model $\mathsf{T}^\theta = \mathsf{H}^\theta \circ \{\mathsf{A}_i^\theta\}_{i=1}^N$ is parameterized by $\theta$, the student model $\mathsf{S}^\psi$ is parameterized by $\psi$, and all additional modules are parameterized by $\phi$, including the prompt blocks $\{\mathsf{P}_i^\phi\}_{i=1}^N$, fusion block $\{\mathsf{F}_i^\phi\}_{i=1}^N$ and initialized classification head $\mathsf{H}^\phi$. During distillation, we optimize the prompt-based forward path through bidirectional supervision from the teacher's original forward path and the student. The former ensures that the performance of the prompt-based forward path gradually increases, while the latter allows the transferred knowledge to be compatible with the representation abilities of the student. Different from other prompt-based tuning strategies, back-propagation in DFPT-KD involves calculating the gradients for all parameters, including $\phi$, $\psi$, and $\theta$. Then, we only fine-tune prompt blocks $\{\mathsf{P}_i^\phi\}_{i=1}^N$, fusion block $\{\mathsf{F}_i^\phi\}_{i=1}^N$ and classification head $\mathsf{H}^\phi$ to optimize the prompt-based forward path. In this way, the gradient information of $\phi$ facilitates the convergence towards the global minimum of the objective function, which is illustrated in Section IV-C4. Formally, the objective of optimization can be formulated as

$$\phi^* = \arg\min_{\phi,\psi} \frac{1}{|\mathcal{D}|} \sum_{j=1}^{|\mathcal{D}|} \lambda \mathcal{L}_{\mathsf{CE}}(p^{\mathsf{P}}, y^j)$$
$$+ (1 - \lambda)\mathsf{L}_{\mathsf{KD}}(p^{\mathsf{P}}, p^{\mathsf{T}}) \quad (11)$$
$$+ (1 - \lambda)\mathsf{L}_{\mathsf{KD}}(p^{\mathsf{P}}, p^{\mathsf{S}}),$$

where $p^{\mathsf{P}} = \mathsf{H}^\phi(\check{\mathsf{x}}_N^j)$, $p^{\mathsf{T}} = \mathsf{H}^\theta(\mathsf{x}_N^j)$, $p^{\mathsf{S}} = \mathsf{S}^\psi(\mathsf{x}^j)$. $\lambda$ is a balanced weight, $\mathsf{L}_{\mathsf{CE}}$ is cross-entropy loss, $\mathsf{L}_{\mathsf{KD}}$ is KL divergence loss.

We can further optimize the whole prompt-based forward path to make the transferred knowledge better compatible with the student's representation abilities. This resulting variant is referred to as DFPT-KD†. Compared to DFPT-KD, DFPT-KD† also fine-tunes the pre-trained teacher backbone $\{\mathsf{A}_i^\theta\}_{i=1}^N$ with a minimal learning rate. As a result, the corresponding objective of optimization can be formulated as

$$\phi^*, \theta^* = \arg\min_{\phi,\psi,\theta} \frac{1}{|\mathcal{D}|} \sum_{j=1}^{|\mathcal{D}|} \lambda \mathcal{L}_{\mathsf{CE}}(p^{\mathsf{P}}, y^j)$$
$$+ (1 - \lambda)\mathsf{L}_{\mathsf{KD}}(p^{\mathsf{P}}, p^{\mathsf{T}}) \quad (12)$$
$$+ (1 - \lambda)\mathsf{L}_{\mathsf{KD}}(p^{\mathsf{P}}, p^{\mathsf{S}}),$$

where $p^{\mathsf{P}}$, $p^{\mathsf{T}}$, and $p^{\mathsf{S}}$ are defined accordingly.

DFPT-KD† improves upon DFPT-KD, boosting the performance of our method to a higher level. That is confirmed in section IV-A.

*C. Distillation with Dual-Forward Path Teacher*

The student's training will be much simpler compared to optimizing the prompt-based forward path. In this work, we focus on exploring how to effectively address the capacity gap problem and promote the student to achieve more comparable performance with the pre-trained teacher. For this purpose, the student is trained to mimic the predicted probabilities of the dual-forward path teacher. In DFPT-KD, the optimization process of the student is presented as

$$\psi^* = \arg\min_{\phi,\psi} \frac{1}{|\mathcal{D}|} \sum_{j=1}^{|\mathcal{D}|} \alpha \mathcal{L}_{\mathsf{CE}}(p^{\mathsf{S}}, y^j)$$
$$+ \beta(\mathsf{L}_{\mathsf{KD}}(p^{\mathsf{S}}, p^{\mathsf{T}}) + \mathsf{L}_{\mathsf{KD}}(p^{\mathsf{S}}, p^{\mathsf{P}})), \quad (13)$$

where $p^{\mathsf{S}}$, $p^{\mathsf{T}}$, and $p^{\mathsf{P}}$ represent the predicted probability of the student, teacher's original forward path (equivalent to the pre-trained teacher), and teacher's prompt-based forward path. $\alpha$ and $\beta$ denote the weight of each loss component. In DFPT-KD†, this optimization process is formulated as

$$\psi^* = \arg\min_{\phi,\psi,\theta} \frac{1}{|\mathcal{D}|} \sum_{j=1}^{|\mathcal{D}|} \alpha \mathcal{L}_{\mathsf{CE}}(p^{\mathsf{S}}, y^j)$$
$$+ \beta(\mathsf{L}_{\mathsf{KD}}(p^{\mathsf{S}}, p^{\mathsf{T}}) + \mathsf{L}_{\mathsf{KD}}(p^{\mathsf{S}}, p^{\mathsf{P}})). \quad (14)$$

The distillation procedure of DFPT-KD and DFPT-KD† are also shown in Algorithm 1.

*D. Theoretical analysis*

The effectiveness of the teacher's prompt-based forward path can be explained based on VC-dimension theory [64]. Generally, a classifier $f_s$ has a classification error that satisfies the following inequality:

$$R(f_s) - R(f_r) \leqslant O\left(\frac{|F_s|_C}{n^{\alpha_{sr}}}\right) + \epsilon_{sr}, \quad (15)$$

where $O(\cdot)$ and $\epsilon$ terms represent estimation and approximation error, respectively. The former is related to the learning process given a group of training data, while the latter reflects the capacity of a network function. Here, $f_r \in F_r$ is the real target function, $f_s \in F_s$ is the student function, $R$ denotes the error, $|\cdot|_C$ is an appropriate capacity measure for function class, $n$ is the number of training data, and finally $\frac{1}{2} \leqslant \alpha \leqslant 1$ is a factor related to the learning rate. Smaller values of $\alpha$

**Algorithm 1:** DFPT-KD and DFPT-KD†

**Input:** Training set $D = \{(x_i, y_i)\}_{i=1}^{|\mathcal{D}|}$, total training epoch $E$, pre-trained teacher $T^\theta$, prompt blocks $\{P_i^\phi\}_{i=1}^N$, fusion block $\{F_i^\phi\}_{i=1}^N$, initialized classification head $H^\phi$, student $S^\psi$;
**Output:** Well-trained student $S^\psi$;

1. Initialize the student, prompt blocks, fusion block, and initialized classification head;
2. Insert $\{P_i^\phi\}_{i=1}^N$, $\{F_i^\phi\}_{i=1}^N$, and $H^\phi$ into the pre-trained teacher to establish the prompt-based forward path, forming the dual-forward path teacher;
3. **for** *epoch* = 1, 2, ..., $E$ **do**
4.    Get a batch of data; Feed the data into the dual-forward path teacher;
5.    Feed the data to the student;
6.    Get the predicted probabilities of the teacher's original forward path, the teacher's prompt-based forward path, and the student;
7.    **if** *using* DFPT-KD **then**
8.       optimizing the prompt-based forward path by Eq. (11);
9.       optimizing the student by Eq. (13);
10.   **end**
11.   **if** *using* DFPT-KD† **then**
12.       optimizing the prompt-based forward path by Eq. (12);
13.       optimizing the student by Eq. (14);
14.   **end**
15.   **return** Student $S^\psi$, prompt block $\{P_i^\phi\}_{i=1}^N$, fusion block $\{F_i^\phi\}_{i=1}^N$, and classification head $H^\phi$;
16. **end**

indicate difficult problems, while larger values correspond to easier problems. Note that $\epsilon_{sr}$ is the approximation error of the best student function concerning the target function. Building on the top of [65], we extend their result and analyze how the knowledge distilled from the prompt-based forward path improves distillation performance. Let $f_t \in F_t$ be the pre-trained teacher function, similar to Eq. (15), we have:

$$R(f_t) - R(f_r) \leq O\left(\frac{|F_t|_C}{n^{\alpha_{tr}}}\right) + \epsilon_{tr}, \quad (16)$$

where $\epsilon_{tr}$ and $\alpha_{tr}$ are defined for the teacher with traditional training approach. After pre-training, we can distill knowledge from the pre-trained teacher to the student:

$$R(f_s) - R(f_t) \leq O\left(\frac{|F_s|_C}{n^{\alpha_{st}}}\right) + \epsilon_{st}, \quad (17)$$

where $\alpha_{st}$ and $\epsilon_{st}$ are related to student learning from a pre-trained teacher. By combining Eq. (16) and (17), it can be seen that vanilla KD has better performance than the traditional training approach if the following inequality holds:

$$O\left(\frac{|F_s|_C}{n^{\alpha_{st}}} + \frac{|F_t|_C}{n^{\alpha_{tr}}}\right) + \epsilon_{st} + \epsilon_{tr} \leq O\left(\frac{|F_s|_C}{n^{\alpha_{sr}}}\right) + \epsilon_{sr}. \quad (18)$$

Lopez et al. [65] pointed out that $|F_t|_C$ should not be large in the finite sample regime. Otherwise, the vanilla KD would be inferior to the traditional training approach. Another distillation failure case happens when the teacher and student have great capacity diversity, which means $\epsilon_{st}$ is very small and even close to $\epsilon_{sr}$. In line with the finding of [65], we also work with the upper bounds, not the actual performance, and in an asymptotic regime. In DFPT-KD and DFPT-KD†, we establish an additional prompt-based forward path within the pre-trained teacher and optimize it through bidirectional supervision from the teacher's original forward path and student:

$$\begin{aligned} R(f_p) - R(f_t) &\leq O\left(\frac{|F_p|_C}{n^{\alpha_{pt}}}\right) + \epsilon_{pt}, \\ R(f_p) - R(f_s) &\leq O\left(\frac{|F_p|_C}{n^{\alpha_{ps}}}\right) + \epsilon_{ps}, \end{aligned} \quad (19)$$

where $f_p \in F_p$ is the prompt-based forward path function, $\alpha_{pt}$, $\epsilon_{pt}$, $\alpha_{ps}$, and $\epsilon_{ps}$ are defined accordingly. Meanwhile, the student mimics the predicted probability of the prompt-based forward path:

$$R(f_s) - R(f_p) \leq O\left(\frac{|F_s|_C}{n^{\alpha_{sp}}}\right) + \epsilon_{sp}. \quad (20)$$

As the prompt-based forward path receives the reversed supervision from the student, their knowledge is compatible with the representation ability of the student and easy to learn, which means $\alpha_{st} \leq \alpha_{sp}$ and $\alpha_{st} \leq \alpha_{ps}$. The significant capacity gap in vanilla KD makes it challenging for the student to mimic the predicted distribution of pre-trained teachers. In our method, the prompt-based forward path is established on top of the original forward path by inserting lightweight modules, minimizing or eliminating the capacity gap between them. Therefore, we have $\alpha_{st} \leq \alpha_{pt}$. Asymptotically speaking, we have $O\left(\frac{|F_s|_C}{n^{\alpha_{sp}}} + \frac{|F_p|_C}{n^{\alpha_{pt}}} + \frac{|F_p|_C}{n^{\alpha_{ps}}}\right) \leq O\left(\frac{|F_s|_C}{n^{\alpha_{st}}}\right)$, which leads to $O\left(\frac{|F_s|_C}{n^{\alpha_{sp}}} + \frac{|F_p|_C}{n^{\alpha_{pt}}} + \frac{|F_p|_C}{n^{\alpha_{ps}}} + \frac{|F_t|_C}{n^{\alpha_{tr}}}\right) \leq O\left(\frac{|F_s|_C}{n^{\alpha_{st}}} + \frac{|F_t|_C}{n^{\alpha_{tr}}}\right)$. Moreover, according to [7], we get $\epsilon_{sp} + \epsilon_{pt} \leq \epsilon_{st}$, which is equivalent to $\epsilon_{sp} + \epsilon_{pt} + \epsilon_{tr} \leq \epsilon_{st} + \epsilon_{tr}$. In most cases, students' capacity is smaller than the teacher's. Therefore, we can infer from [66] that $\epsilon_{ps} = 0$. As a result, we can conclude the following inequality:

$$\begin{aligned} & O\left(\frac{|F_s|_C}{n^{\alpha_{sp}}} + \frac{|F_p|_C}{n^{\alpha_{pt}}} + \frac{|F_p|_C}{n^{\alpha_{ps}}} + \frac{|F_t|_C}{n^{\alpha_{tr}}}\right) + \epsilon_{sp} + \epsilon_{pt} + \epsilon_{ps} + \epsilon_{tr} \\ & \leq O\left(\frac{|F_s|_C}{n^{\alpha_{st}}} + \frac{|F_t|_C}{n^{\alpha_{tr}}}\right) + \epsilon_{st} + \epsilon_{tr}. \end{aligned} \quad (21)$$

It can be inferred from Eq. (21) that the student performs better when learning from the teacher's prompt-based forward path than from the original forward path, demonstrating the effectiveness of our approach.

## IV. EXPERIMENTS

We conduct experiments on various neural networks, such as VGG [3], ResNet [2], Wide ResNet [67] (abbreviated as WRN), ShuffleNetV1 [68], ShuffleNetV2 [69] and MobileNetV2 [70]. In this section, we first describe the experimental settings used for experiments. Then, we evaluate the performance of our method on CIFAR-100 [35], ImageNet [36], and CUB-200 [37]. After that, extensive analyses are conducted to investigate critical components of our method.

*A. Datasets and Experimental Details*

*1) CIFAR-100:* This dataset is a medium-scale image classification dataset consisting of 60k images (50,000 training samples and 10,000 testing samples) from 100 categories. The resolution of each image is $32 \times 32$. In the distillation process, we set batch size as 64 and base learning rate as 0.05 (0.01 for ShuffleNetV1, ShuffleNetV2, and MobileNetV2), which is divided by 10 at 150, 180, and 210 epochs, respectively. We use the SGD optimizer [71] and take 1 GPU to take experiments. All results are the average of five trials.

*2) ImageNet:* This dataset is a large-scale benchmark dataset for image classification, with a total of 1.28 million training samples and 50,000 testing samples from 1000 categories. The resolution of input samples is fixed to $224 \times 224$. In the distillation process, we set batch size as 512 and base learning rate as 0.1, with a decay rate of 0.1 at the 30, 60, and 90 epochs, respectively. We use the SGD optimizer and take 4 GPUs to take experiments. All results are the average of three trials.

*3) CUB-200:* This dataset is a common fine-grain classification benchmark dataset consisting of 11,788 images (5,994 training samples and 5,794 testing samples) across 200 bird subcategories. The resolution of input samples is fixed to $224 \times 224$. In the distillation process, we set batch size as 64 and the learning rate as 0.05 for VGG, ResNet, and WRN, and 0.01 for ShuffleNetV1/V2 and MobileNetV2, which is reduced by a rate of 0.1 at 150, 180, and 210 epochs, respectively. We use the SGD optimizer and take 2 GPUs to conduct experiments. All results are the average of five trials.

*B. Experimental Results*

*1) Results on CIFAR-100:* We evaluate the performance of our method on CIFAR-100 and compare it with ten feature distillation methods, FitNet [13], AT [14], RKD [43], VID [45], CRD [42], WCoRD [16], CoCoRD [47], NROM [48], ReviewKD [15], CAT-KD [19], and nine logit distillation methods, KD [7], DML [8], DKD [9], KCKD [17], CTKD [10], MLD [11], SD-KD [40], WTTM [41], and LSKD [12]. For the homogeneous teacher-student pairs, as shown in Tab. I, DFPT-KD obtains 2.26% - 4.98% absolute gains than baseline and outperforms KD with 0.66% - 4.15% margins. DFPT-KDt obtains 3.40% - 6.13% absolute gains than baseline and outperforms KD with 1.57% - 5.30% margins. When it comes to the experiment where the teacher and student are in heterogeneous architecture, the results in Tab. II show that DFPT-KD achieves more significant gains with 5.66% - 7.22% margins than baseline and surpass KD with 2.70% - 4.29% gains. DFPT-KDt obtains 6.22% - 8.55% absolute gains than baseline and outperforms KD with 3.45% - 5.40% margins. In general, DFPT-KD and DFPT-KDt always outperform KD, consistent with the theoretic analysis in Section III-D. Additionally, DFPT-KDt is better than DFPT-KD in distillation performance and achieves new state-of-the-art results. These results validate the effectiveness of the proposed method in addressing the capacity gap problem. Moreover, DFPT-KDt even boosts the student to perform better than the pre-trained teacher. For example, when the teacher-student pair is WRN-40-2 → ShuffleNetV1, DFPT-KDt obtains 78.28% Top-1 accuracy on the CIFAR-100, surpassing teacher by 2.67%.

*2) Results on ImageNet:* To further verify the effectiveness of the proposed method, we conduct experiments on the more challenging ImageNet dataset. We compare DFPT-KD and DFPT-KDt with eleven state-of-the-art distillation methods, including KD [7], AT [14], CRD [42], SRRL [73], ReviewKD [15], MGD [74], DKD [9], MLD [11], CAT-KD [19], SD-KD [40], LSKD [12]. As shown in Tab. III, DFPT-KD consistently improves Top-1 and Top-5 accuracies over KD, achieving comparable or even better performance with other state-of-the-art methods. DFPT-KDt improves upon DFPT-KD and achieves new state-of-the-art results. Specifically, when the teacher-student pair is ResNet34 → ResNet18, DFPT-KDt obtains 1.77% Top-1 and 0.90% Top-5 gains over KD. When the teacher-student pair is ResNet50 → MobileNetV2, DFPT-KDt brings 2.96% Top-1 and 1.33% Top-5 accuracy improvements over KD. These results demonstrate that our method is still effective on the large-scale dataset.

*3) Results on CUB-200:* In addition to the commonly evaluated classification datasets (*i.e.* CIFAR-100 and ImageNet), we also compared our method with KD [7], SP [44], CRD [42], SemCKD [75], ReviewKD [15], DKD [9], NKD [76], and SD-KD [40] on CUB-200. In this fine-grained classification task, different classes have small discrepancies. The results in Tab. IV show that the proposed method is superior to other KD methods in most cases. Specifically, DFPT-KD and DFPT-KDt outperform KD with 3.55% - 10.41% and 4.41% - 11.48% margins, respectively. The reason may be that fine-grained classification tasks have a stronger demand for fine-grained feature discriminatory capacity than conventional classification tasks since different classes have similar feature information. In the proposed method, the student can learn accurate and compatible knowledge representations from the dual-forward path teacher simultaneously, bridging the capacity gap and promoting the student to achieve more comparable performance with the pre-trained teacher. That demonstrates the potential of our method for distilling fine-grained classification models.

*C. Analyses*

*1) The Capacity Gap Problem:* It is interesting to explore the performance gap between the pre-trained teacher and student in our method. As shown in Tab. V, we calculate the gap between the accuracies of the pre-trained teacher and students. When the student outperforms the teacher, the corresponding gap is negative. It can be observed that with our DFPT-KD, the student achieves more comparable performance with the pre-trained teacher than KD. DFPT-KDt improves upon DFPT-KD and makes the student even outperform the pre-trained teacher in some cases. This is because fine-tuning the whole prompt-based forward path makes the transferred knowledge better compatible with the representation abilities of the student, which is beneficial to bridge the capacity gap.

TABLE I
DISTILLATION PERFORMANCE COMPARISON BETWEEN HOMOGENEOUS ARCHITECTURE. IT REPORTS TOP-1 ACCURACY (%) ON CIFAR-100 VALIDATION SET. THE BEST AND THE SECOND-BEST RESULTS ARE INDICATED IN **BOLD** AND <u>UNDERLINE</u>.

| Type | | ResNet32×4 | ResNet110 | ResNet56 | WRN-40-2 | WRN-40-2 | VGG13 |
|---|---|---|---|---|---|---|---|
| Type | Teacher | ResNet32×4 | ResNet110 | ResNet56 | WRN-40-2 | WRN-40-2 | VGG13 |
| | Acc | 79.42 | 74.31 | 72.34 | 75.61 | 75.61 | 74.64 |
| | Student | ResNet8×4 | ResNet32 | ResNet20 | WRN-40-1 | WRN-16-2 | VGG8 |
| | Acc | 72.50 | 71.14 | 69.06 | 71.98 | 73.26 | 70.36 |
| Feature | FitNet [13] | 73.50 | 71.06 | 69.21 | 72.24 | 73.58 | 71.02 |
| | AT [14] | 73.44 | 70.55 | 72.77 | 74.08 | 71.43 | |
| | RKD [43] | 71.90 | 71.82 | 69.61 | 72.22 | 73.35 | 71.48 |
| | VID [45] | 73.09 | 72.61 | 70.38 | 73.30 | 74.11 | 71.23 |
| | CRD [42] | 75.51 | 73.48 | 71.16 | 74.14 | 75.48 | 73.94 |
| | ReviewKD [15] | 75.63 | 73.89 | 71.89 | 75.09 | 76.12 | 74.84 |
| | WCoRD [16] | 75.95 | 73.81 | 71.56 | 74.73 | 75.88 | 74.55 |
| | CoCoRD [47] | 75.29 | 74.10 | 71.74 | 75.17 | 75.48 | 73.99 |
| | NORM [48] | 76.49 | 73.67 | 71.35 | 74.82 | 75.65 | 73.95 |
| | CAT-KD [19] | 76.91 | 73.62 | 71.62 | 74.82 | 75.60 | 74.65 |
| Logit | KD [7] | 73.33 | 73.08 | 70.66 | 73.54 | 74.92 | 72.98 |
| | DML [8] | 72.12 | 72.03 | 69.52 | 72.68 | 73.58 | 71.79 |
| | DKD [9] | 76.32 | 74.11 | 71.97 | 74.81 | 76.24 | 74.68 |
| | KCKD [72] | 74.05 | N/A | 69.52 | 72.68 | 73.58 | 71.79 |
| | CTKD [10] | N/A | 73.52 | 71.19 | 73.93 | 75.45 | 73.52 |
| | MLD [11] | 77.08 | 74.11 | <u>72.19</u> | 75.35 | 76.63 | 75.18 |
| | SD-KD [40] | 75.58 | 73.57 | 71.20 | 74.52 | 76.09 | 74.18 |
| | WTTM [41] | 76.06 | 74.13 | 71.92 | 74.58 | 76.37 | 74.44 |
| | LSKD [12] | 76.62 | <u>74.17</u> | 71.43 | 74.37 | 76.11 | 74.36 |
| Ours | DFPT-KD | <u>77.48</u> | 73.92 | 71.32 | <u>75.47</u> | <u>76.68</u> | <u>75.23</u> |
| | DFPT-KD† | **78.63** | **74.65** | **72.46** | **76.38** | **77.23** | **75.74** |

We also compare our method with some typical adaptive KD methods, including RCO [28], TAKD [21], DGKD [23], and AAKD [27], on CIFAR-100. As shown in Tab. VI, DFPT-KD and DFPT-KD† consistently perform better than these methods, indicating the superiority of our method in alleviating the capacity gap problem. In addition to the superior performance, as shown in Tab. VII, the parameters and floating-point operations (FLOPs) of the prompt blocks and fusion blocks are lower than that of the TA in TAKD, DGKD and AAKD, consuming less storage and computation resource.

*2) The Guidance of Dual-Forward Path:* To further analyse the guidance of the dual-forward path, we let the student exclusively learn knowledge from the prompt-based forward path or original forward path. The experiments are conducted on CIFAR-100, and the teacher-student pair is ResNet32×4 → ResNet8×4. As shown in Fig. 3 (a), learning from the prompt-based forward path achieves better performance than the original forward path, validating its effectiveness in bridging the capacity gap. Moreover, learning from the dual-forward path achieves optimal distillation performance, consistent with our viewpoint that accurate and compatible knowledge representations are indispensable for the distillation process.

*3) The Bidirectional Supervision:* In the proposed method, we optimize the prompt-based forward path through a bidirectional supervision approach. The distillation loss function includes the teacher's original forward path supervision $L_{KD}(p^P, p^T)$, and student's reversed supervision $L_{KD}(p^P, p^S)$. We conduct related experiments to explore the influence of these two terms in Eq. (11). The results are shown in the Fig. 3 (b). It is evident that the teacher's original forward path supervision $L_{KD}(p^P, p^T)$ has a promotion effect, making the performance of prompt-based forward path increased. In contrast, the student's reversed supervision $L_{KD}(p^P, p^S)$ makes the prompt-based forward path can be optimized by the student. Thus, the bidirectional supervision approach makes the knowledge of prompt-based forward path follow an easy-to-hard principle and compatible with the representation abilities of the student at each distillation phase.

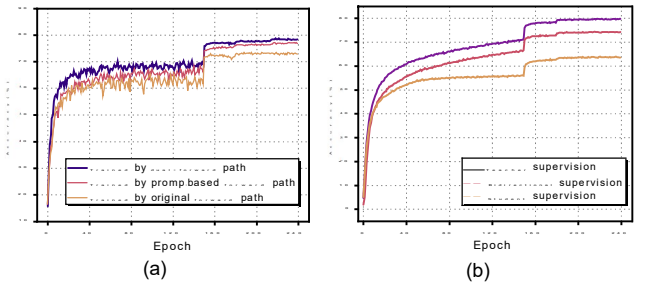

Fig. 3. (a) The validation accuracy curves of the student under three kinds of guidance approaches. (b) The training accuracy curves of the prompt-based forward path under three kinds of supervision approaches.

To further illustrate that the knowledge of prompt-based forward path is more conducive to student learning, we show the similarity between students and dual-forward path teachers. We employ KL divergence as the similarity metric [20], where lower KL divergence implies higher similarity. Fig. 4 (a) and (c) present the similarities between the outputs of students and dual-forward path teachers. Fig. 4 (b) and (d) present the comparison of 1 - $p_t^T$ and 1 - $p_t^P$. These results show that learning from the prompt-based forward path is more conducive to the student than the original forward path (equivalent to the pre-trained teacher), bridging the capacity gap.

TABLE II
DISTILLATION PERFORMANCE COMPARISON BETWEEN HETEROGENEOUS ARCHITECTURE. IT REPORTS TOP-1 ACCURACY (%) ON CIFAR-100 VALIDATION SET. THE BEST AND THE SECOND-BEST RESULTS ARE INDICATED IN **BOLD** AND <u>UNDERLINE</u>.

| Type | | ResNet32×4 | WRN-40-2 | VGG13 | ResNet50 | ResNet32×4 |
|---|---|---|---|---|---|---|
| | Teacher | | | | | |
| | Acc | 79.42 | 75.61 | 74.64 | 79.34 | 79.42 |
| | Student | ShuffleNetV1 | ShuffleNetV1 | MobileNetV2 | MobileNetV2 | ShuffleNetV2 |
| | Acc | 70.50 | 70.50 | 64.60 | 64.60 | 71.82 |
| Feature | FitNet [13] | 73.59 | 73.73 | 64.14 | 63.16 | 73.54 |
| | AT [14] | 71.73 | 73.32 | 59.40 | 58.58 | 72.73 |
| | RKD [43] | 72.28 | 72.21 | 64.52 | 64.43 | 73.21 |
| | VID [45] | 73.38 | 73.61 | 65.56 | 67.57 | 73.40 |
| | CRD [42] | 75.11 | 76.05 | 69.73 | 69.11 | 75.65 |
| | ReviewKD [15] | 77.45 | 77.14 | 70.37 | 69.89 | 77.78 |
| | WCoRD [16] | 75.40 | 76.32 | 69.47 | 70.45 | 75.96 |
| | CoCoRD [47] | 75.99 | 76.42 | 69.86 | 70.22 | 77.28 |
| | NORM [48] | 77.42 | 77.06 | 68.94 | 70.56 | 78.07 |
| | CAT-KD [19] | 78.26 | 77.35 | 69.13 | 71.36 | 78.41 |
| Logit | KD [7] | 74.07 | 74.83 | 67.37 | 67.35 | 74.45 |
| | DML [8] | 72.89 | 72.76 | 65.63 | 65.71 | 73.45 |
| | DKD [9] | 76.45 | 76.70 | 69.71 | 70.35 | 77.07 |
| | KCKD [17] | 74.33 | 75.60 | 68.61 | 67.94 | 75.19 |
| | CTKD [10] | 74.48 | 75.78 | 68.46 | 68.47 | 75.31 |
| | MLD [11] | 77.18 | 77.44 | <u>70.57</u> | <u>71.04</u> | 78.44 |
| | SD-KD [40] | 76.30 | 76.65 | 68.79 | 69.55 | 76.67 |
| | WTTM [41] | 74.37 | 75.42 | 69.16 | 69.59 | 76.55 |
| | LSKD [12] | N/A | N/A | 68.61 | 69.02 | 75.56 |
| Ours | DFPT-KD | <u>77.72</u> | <u>77.53</u> | 70.26 | 70.89 | <u>78.74</u> |
| | DFPT-KD† | **79.05** | **78.29** | **70.82** | **71.73** | **79.85** |

TABLE III
DISTILLATION PERFORMANCE COMPARISON BETWEEN HOMOGENEOUS AND HETEROGENEOUS ARCHITECTURES. IT REPORTS TOP-1 AND TOP-5 ACCURACIES (%) ON THE IMAGENET VALIDATION SET. THE BEST AND THE SECOND-BEST RESULTS ARE INDICATED IN **BOLD** AND <u>UNDERLINE</u>.

| | | Top-1 | Top-5 | Top-1 | Top-5 |
|---|---|---|---|---|---|
| Type | Teacher | ResNet34 | | ResNet50 | |
| | | 73.31 | 91.42 | 76.16 | 92.86 |
| | Student | ResNet18 | | MobileNetV2 | |
| | | 69.75 | 89.07 | 68.87 | 88.76 |
| Feature | AT [14] | 70.69 | 90.01 | 69.56 | 89.33 |
| | CRD [42] | 71.17 | 90.13 | 71.37 | 90.41 |
| | SRRL [73] | 71.73 | 90.60 | 72.49 | 90.92 |
| | ReviewKD [15] | 71.61 | 90.51 | 72.56 | 91.00 |
| | MGD [74] | 71.58 | 90.35 | 72.35 | 90.71 |
| | CAT-KD [19] | 71.26 | 90.45 | 72.24 | 91.13 |
| Logit | KD [7] | 70.66 | 89.88 | 70.50 | 90.34 |
| | DKD [9] | 71.70 | 90.41 | 72.05 | 91.05 |
| | MLD [11] | 71.90 | 90.55 | 73.01 | <u>91.42</u> |
| | SD-KD [40] | 71.44 | 90.05 | 72.24 | 90.71 |
| | LSKD [12] | 71.42 | 90.29 | 72.18 | 90.80 |
| Ours | DFPT-KD | <u>72.08</u> | <u>90.62</u> | <u>73.15</u> | 91.34 |
| | DFPT-KD† | **72.43** | **90.78** | **73.46** | **91.67** |

*4) The Optimization of Prompt-Based Forward Path:* When optimizing the prompt-based forward path, the main difference between our approach and other prompt-based tuning strategies is that we calculate the gradients for all parameters. That is reasonable since we want to dynamically adjust the transferred knowledge of the prompt-based forward path rather than that of the prompt blocks and fusion blocks. To better illustrate this, we take an experiment on CIFAR-100 and show the training accuracy and loss curves of the prompt-based forward path in DFPT-KD, DFPT-KD†, and base approach (not calculating the gradients of pre-trained teacher). As shown in Fig. 5 (a), the training accuracy in DFPT-KD and DFPT-KD† gradually increased to provide the student with an easy-to-hard and compatible knowledge sequence. In contrast, the accuracy curve in the base approach exhibits a slower increasing rate and stabilizes at lower values. In the meantime, we can observe from Fig. 5 (b) that the loss curve in the base approach decreases slowly and retains relatively high and fluctuating values. In contrast, the loss curve in DFPT-KD and DFPT-KD† decreases more consistently, with DFPT-KD† showing the lowest converge loss (4.81). These adequately demonstrate the rationality of the proposed optimization approach for the prompt-based forward path.

TABLE IV
DISTILLATION PERFORMANCE COMPARISON ACROSS THREE DIFFERENT TEACHER-STUDENT PAIRS. IT REPORTS TOP-1 ACCURACY (%) ON THE CUB-200 VALIDATION SET. THE BEST AND THE SECOND-BEST RESULTS ARE INDICATED IN **BOLD** AND <u>UNDERLINE</u>.

| Type | | ResNet32×4 66.17 MobileNetV2 40.23 | ResNet32×4 66.17 ShuffleNetV1 37.28 | VGG13 70.19 MobileNetV2 40.23 | VGG13 70.19 VGG8 46.32 | ResNet50 60.01 ShuffleNetV1 37.28 |
|---|---|---|---|---|---|---|
| | Teacher Acc Student Acc | | | | | |
| Feature | SP [44] | 48.49 | 61.83 | 44.28 | 54.78 | 55.31 |
| | CRD [42] | 57.45 | 62.28 | 56.45 | 66.10 | 57.45 |
| | SemCKD [75] | 56.89 | 63.78 | **68.23** | 66.54 | 57.20 |
| | ReviewKD [15] | N/A | 64.12 | 58.66 | 67.10 | N/A |
| | MGD [74] | N/A | N/A | N/A | 66.89 | 57.12 |
| Logit | KD [7] | 56.09 | 61.68 | 53.98 | 64.18 | 57.21 |
| | NKD [76] | 59.81 | 64.00 | 58.40 | 67.16 | 59.11 |
| | DKD [9] | 59.94 | 64.51 | 58.45 | 67.20 | 59.21 |
| | SD-KD [40] | 60.51 | 65.46 | 59.80 | 67.32 | 60.56 |
| Ours | DFPT-KD | 61.26 | 66.32 | 64.39 | 67.73 | 62.48 |
| | DFPT-KD† | **62.03** | **67.15** | <u>65.46</u> | **68.59** | **63.65** |

TABLE V
PERFORMANCE GAP BETWEEN TEACHER AND STUDENT. WE CONDUCTED EXPERIMENTS ON CIFAR-100. NOTE THAT THE GAP IS NEGATIVE WHEN THE STUDENT OUTPERFORMS THE TEACHER. THE SETTINGS ARE THE SAME AS TAB. I AND II.

| Teacher | 79.42 | 74.31 | 72.34 | 75.61 | 75.61 | 74.64 | 79.42 | 75.61 | 74.64 | 79.34 | 79.42 | 76.39 (Avg) |
|---|---|---|---|---|---|---|---|---|---|---|---|---|
| Student | 72.50 | 71.14 | 69.06 | 71.98 | 73.26 | 70.36 | 70.50 | 70.50 | 64.60 | 64.60 | 71.82 | 70.03 (Avg) |
| Gap (Base) | 6.92 | 3.17 | 3.28 | 3.63 | 2.35 | 4.28 | 8.92 | 5.11 | 10.04 | 14.74 | 7.60 | 6.37 (Avg) |
| Gap (KD) | 6.09 | 1.23 | 1.68 | 2.07 | 0.69 | 1.66 | 5.35 | 0.78 | 7.27 | 11.99 | 4.97 | 3.98 (Avg) |
| Gap (DFPT-KD) | 1.94 | 0.39 | 1.02 | 0.14 | -0.85 | -0.49 | 1.70 | -1.92 | 4.38 | 8.45 | 0.68 | 1.41 (Avg) |
| Gap (DFPT-KD†) | 0.79 | -0.34 | -0.12 | -0.77 | -1.62 | -1.10 | 0.37 | -2.68 | 3.83 | 7.61 | -0.43 | 0.50 (Avg) |

*D. Ablation Study of Prompt Blocks and Fusion Blocks*

In this subsection, we perform extensive ablation studies on the CIFAR-100 dataset to systematically analyze the proposed prompt blocks and fusion blocks. If not stated particularly, the teacher-student pair is set to ResNet32×4 → ResNet8×4, and the experimental setting is the same as Section IV-A.

*1) Influence of $r_1$ and $r_2$:* In the prompt blocks, we use channel down-sampling and partial convolution to reduce the storage and computation costs. To reveal the influence of these two operations, we show the validation accuracy, required parameters, and FLOPs under different down-sampling rates $r_1$ and partial convolution ratio $r_2$ settings. As shown in Tab. VIII, a higher down-sampling rate and lower partial convolution ratio consume less storage and computation resources. While too-high values of $r_1$ or too-small values of $r_2$ would render prompt blocks lacking enough spatial representation ability, making them less effective in adjusting the intermediate features of pre-trained teachers. In our experiment, we set $r_1 = \{4, 4, 4, 4\}$ and $r_2 = 0.50$, achieving a better trade-off between training efficiency and distillation performance.

*2) Investigation of inserting approaches:* We also explore the impact of different inserting positions of prompt blocks. Concretely, we compare DFPT-KD and DFPT-KD† with their variants that append all prompt blocks after a single position of network stages. For a fair comparison, these inserting strategies have almost the same complexity as our adopted multi-stage approach. As shown in Fig. 6 (a), we can observe that the multi-stage inserting approach performs better than the single-stage approach, with the 0.65%-0.99% (DFPT-KD) and 0.77%-1.34% (DFPT-KD†) validation accuracy gains. Interestingly, one can observe in Fig. 6 (b) that inserting after a single position quickly improves the performance of prompt-based forward path to saturated, failing to effectively adjust the transferred knowledge. The reason behind this phenomenon is that the proposed multi-stage inserting approach can adjust the low-level, mid-level, and high-level feature representations, which is beneficial for optimizing the Eq. (11) and (12), and make the transferred knowledge can be dynamically adjusted.

*3) Investigation to the capacity of prompt blocks:* To investigate the impact of the capacity of prompt blocks, we compare DFPT-KD and DFPT-KD† with their variants that append multiple prompt blocks to each stage of pre-trained teacher models. The results in Fig. 6 (c) show that increasing the capacity of prompt blocks does not bring significant performance gains to distillation, causing unnecessary computation and storage costs. That indicates the adopted approach in our method suffices to effectively adjust the transferred knowledge using only one prompt block appended to each stage of pre-trained teachers and improve distillation performance.

TABLE VI
DISTILLATION PERFORMANCE COMPARED WITH ADAPTIVE KD METHODS. IT REPORTS TOP-1 ACCURACY (%) ON CIFAR-100 VALIDATION SET. THE BEST AND THE SECOND-BEST RESULTS ARE INDICATED IN **BOLD** AND <u>UNDERLINE</u>.

| Teacher | ResNet32×4 | WRN-40-2 | WRN-40-2 | ResNet32×4 | ResNet32×4 | ResNet50 |
|---|---|---|---|---|---|---|
| Acc | 79.42 | 75.61 | 75.61 | 79.42 | 79.42 | 79.34 |
| Student | ResNet8×4 | WRN-40-1 | WRN-16-2 | ShuffleNetV1 | ShuffleNetV2 | VGG8 |
| Acc | 72.50 | 71.98 | 73.26 | 70.50 | 71.82 | 70.36 |
| TA | ResNet14×4 | WRN-22-2 | WRN-22-2 | ResNet14×4 | ResNet14×4 | ResNet18 |
| Acc | 76.17 | 74.92 | 74.92 | 76.17 | 76.17 | 73.79 |
| RCO [28] | 74.63 | 74.15 | 75.28 | 75.31 | 75.12 | 74.89 |
| TAKD [21] | 74.91 | 73.99 | 75.62 | 74.93 | 75.88 | 74.51 |
| DGKD [23] | 75.01 | 74.33 | 76.10 | 76.13 | 76.41 | 75.21 |
| AAKD [27] | 75.01 | 74.21 | 75.66 | 75.20 | 75.98 | 74.51 |
| DFPT-KD | <u>77.48</u> | <u>75.47</u> | <u>76.46</u> | <u>77.72</u> | <u>78.74</u> | <u>76.39</u> |
| DFPT-KD† | **78.63** | **76.38** | **77.23** | **79.05** | **79.85** | **77.26** |

TABLE VII
COMPARISON OF THE PARAMETERS AND FLOPS BETWEEN THE PROMPT BLOCKS AND FUSION BLOCKS IN EACH PRE-TRAINED TEACHER AND CORRESPONDING TEACHER ASSISTANTS FOR ONE INPUT IMAGE.

| Model | Parameters | FLOPs |
|---|---|---|
| WRN-22-2 | 1.09M | 159.41M |
| ResNet14×4 | 2.78M | 405.99M |
| ResNet18 | 11.22M | 557.94M |
| Prompt Blocks (WRN-40-2) | 0.03M | 5.87M |
| Prompt Blocks (ResNet32×4) | 0.19M | 31.07M |
| Prompt Blocks (ResNet50) | 0.57M | 68.06M |

TABLE VIII
COMPARISON OF THE TOP-1 VALIDATION ACCURACY, PARAMETERS, AND FLOPS UNDER DIFFERENT DOWN-SAMPLING RATES $r_1$ OR PARTIAL CONVOLUTION RATIO $r_2$ SETTINGS. THE COMPUTATION APPROACH OF PARAMETERS AND FLOPS IS THE SAME AS TAB. VII.

| $r_1$ | $r_2$ | Parameters | FLOPs | Accuracy |
|---|---|---|---|---|
| {1, 1, 1, 1} | 1.00 | 3.57M | 561.27M | 78.17 |
| | 0.50 | 1.29M | 203.45M | 78.04 |
| | 0.25 | 0.72M | 113.99M | 78.21 |
| {2, 2, 2, 2} | 1.00 | 1.02M | 157.96M | 78.27 |
| | 0.50 | 0.43M | 68.51M | 78.75 |
| | 0.25 | 0.29M | 46.14M | 78.46 |
| {4, 4, 4, 4} | 1.00 | 0.34M | 53.43M | 78.24 |
| | 0.50 | 0.19M | 31.07M | 78.63 |
| | 0.25 | 0.16M | 25.48M | 78.45 |
| {2, 4, 6, 8} | 1.00 | 0.19M | 46.57M | 78.07 |
| | 0.50 | 0.14M | 28.03M | 77.91 |
| | 0.25 | 0.12M | 23.49M | 77.74 |

*4) Investigation to the multi-scale partial convolution:*
To evaluate the effect of multi-scale partial convolution in prompt blocks, we compare DFPT-KD and DFPT-KDt with its variants that replace the multi-scale partial convolution with ordinary single-scale ones (*e.g.*, 3×3 partial convolution). In addition, we also evaluate the effect of different kernel sizes. As shown in Fig. 6 (d), when adopting single-scale partial convolution, the distillation performance generally improves with the increase of kernel size. Then, it reaches saturation or decreases once the kernel size exceeds a certain value. The reason may be that increasing the receptive field size promotes the representation ability of a neural network. However, too large kernel sizes could make optimizing prompt blocks difficult [77], limiting the performance gains. This similar phenomenon has also been observed when we adopt multi-scale partial convolution with large kernel sizes. Moreover, using ordinary single-scale convolution is inferior to multi-scale ones. That is reasonable since multi-scale convolution can extract feature information at multi-scale levels where smaller kernels capture local details while larger kernels capture broader structural information. In the prompt blocks, we use multi-scale partial convolution to dynamically adjust the intermediate features of pre-trained teachers, allowing the transferred knowledge to be better compatible with the representation ability of the student. Thus, we set the multi-scale kernel size as [3×3, 5×5, 7×7] in our experiments unless specified otherwise.

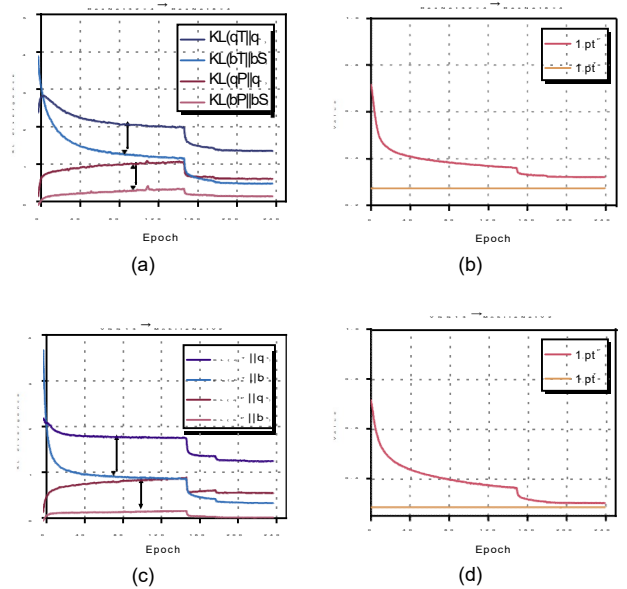

Fig. 4. Teacher-student pairs are ResNet32×4 → ResNet8×4 in (a) and (b), and VGG13 → MobileNetV2 in (c) and (d). We show a comparison of KL divergence among the outputs of the student, teacher's original forward path, and teacher's prompt-based forward path in (a) and (c), and a comparison of $1 - p_t^T$ and $1 - p_t^P$ in (b) and (d).

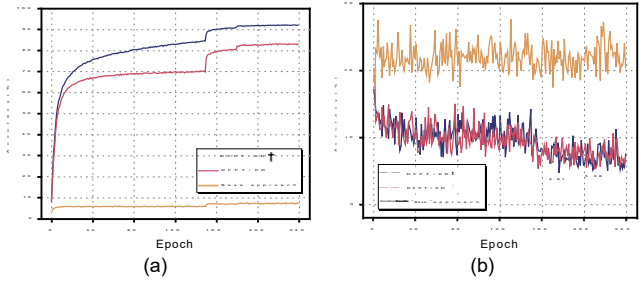

Fig. 5. (a) The training accuracy curves of the prompt-based forward path in DFPT-KD, DFPT-KDt, and base approach. (b) The loss curves of the prompt-based forward path in DFPT-KD, DFPT-KDt, and base approach.

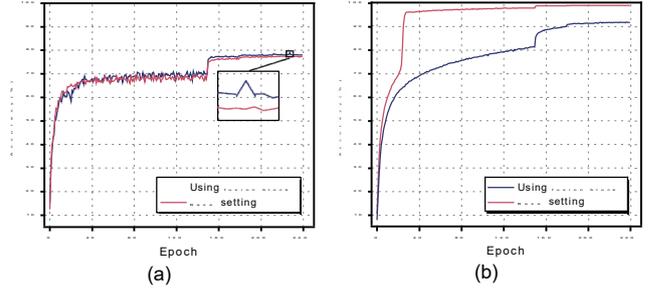

Fig. 7. (a) The validation accuracy curves of the student (DFPT-KDt) under using fusion block or base setting. (b) The training accuracy curves of the prompt-based forward path (DFPT-KDt) under using fusion block or base setting.

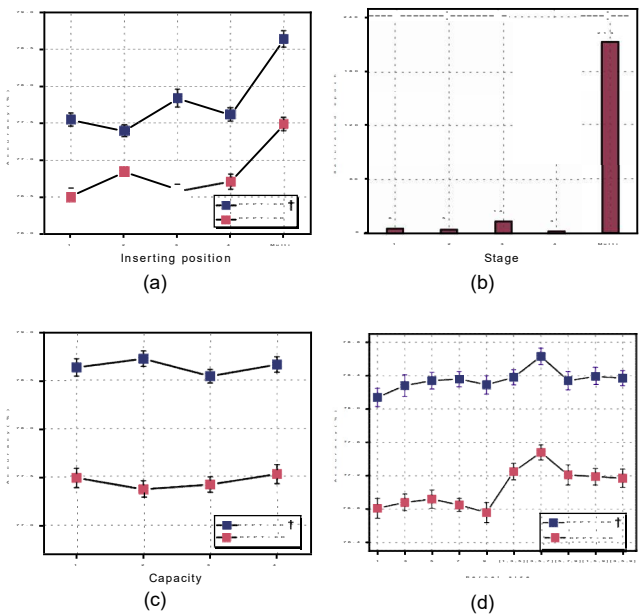

Fig. 6. (a) Validation accuracy of our method under different prompt block inserting approaches. (b) The saturated epoch of prompt-based forward path under different inserting approaches. (c) Validation accuracy of our method under different capacities of prompt blocks. (d) Validation accuracy of our method under different convolution settings.

*5) Impact of the fusion block:* To evaluate the effect of the fusion block, we compare DFPT-KDt with its variant, which removes the fusion block and adds the learned prompts to the corresponding features (base setting). As shown in Fig 7 (a), using the fusion block achieves superior performance, outperforming the base setting. That adequately validates the effectiveness of the fusion block. In addition, we can observe in Fig 7 (b) that fusion block can significantly delay the saturated epoch, making the prompt-based forward path provide the student with an easy-to-hard and compatible knowledge sequence. A considerable reason is that the fusion block can automatically learn the weights for the intermediate features and learned prompts through optimization, automatically adjusting the relationship between these two terms.

## V. CONCLUSION AND FUTURE WORK

In this paper, we present **D**ual-**F**orward **P**ath **T**eacher **K**nowledge **D**istillation (DFPT-KD), which replace the pre-trained teacher with a novel dual-forward path teacher to supervise the learning of student, effectively addressing the capacity gap problem and promoting the student to achieve more comparable performance with the pre-trained teacher. The main idea of DFPT-KD is establishing an additional prompt-based forward path within the pre-trained teacher and optimizing it with the pre-trained teacher frozen to make the transferred knowledge dynamically adjusted to be compatible with the representation abilities of the student. Furthermore, a more powerful version of DFPT-KD, dubbed DFPT-KDt, is proposed to achieve more significant improvements. Extensive experiments on three image classification datasets demonstrate the effectiveness of our method.

We believe the proposed dual-forward path teacher and corresponding learning approach are promising in the knowledge distillation community. In the future, we plan to explore the potential of our approach in combination with other distillation approaches and apply it to more practical tasks.

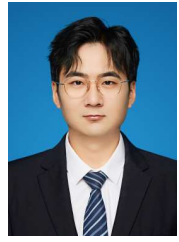

**Tong Li** received B.S. degree from Xi'an University of Technology, Xi'an, China, in 2019. He is currently pursuing a Ph.D. degree in control theory and control engineering at the Xi'an University of Technology, Xi'an, China. His research interests include machine learning, knowledge distillation and computer vision.

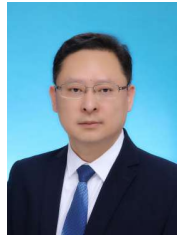

**Long Liu** received the Ph.D. degree from Xi'an Jiaotong University in 2005. He has attended the School of Computer Science of Carnegie Mellon University in the United States for research. He is currently a professor in the School of Automation and Information Engineering at Xi'an University of Technology. His research interests include vision action recognition, vision tracking and detection, and model compression.

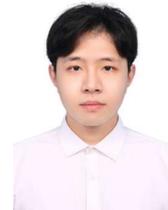

**Yihang Hu** is pursuing a master's degree at the School of Automation and Information Engineering, Xi'an University of Technology, Xi'an, China. His research interests include computer vision, knowledge distillation, and model compression.

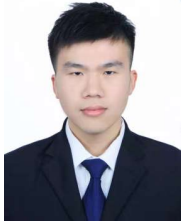

**Hu Chen** is pursuing an M.S. degree at the School of Automation and Information Engineering, Xi'an University of Technology, China. His research interests lie in machine Learning and visual tracking.

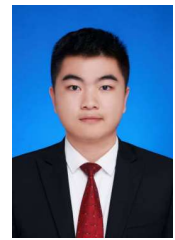

**Shifeng Chen** is pursuing an M.S. degree at the School of Automation and Information Engineering, Xi'an University of Technology, China. His research interests lie in machine Learning and image classification.